\documentclass[10pt,twocolumn,letterpaper]{article}

\usepackage{cvpr}
\usepackage{times}
\usepackage{epsfig}
\usepackage{graphicx}
\usepackage{amsmath}
\usepackage{amssymb}
\usepackage{adjustbox}
\usepackage{mathtools}
\DeclarePairedDelimiter{\floor}{\lfloor}{\rfloor}

\usepackage[pagebackref=true,breaklinks=true,letterpaper=true,colorlinks,bookmarks=false]{hyperref}

\cvprfinalcopy % *** Uncomment this line for the final submission

\def\cvprPaperID{2136} % *** Enter the CVPR Paper ID here

\ifcvprfinal\fi

\begin{document}

%%%%%%%%% TITLE
\title{3D Shape Induction from 2D Views of Multiple Objects}

\author{Matheus Gadelha, Subhransu Maji and Rui Wang\\
University of Massachusetts - Amherst\\
{\tt\small \{mgadelha, smaji, ruiwang\}@cs.umass.edu}
}

\maketitle

\begin{abstract}
In this paper we investigate the problem of inducing a distribution over three-dimensional structures given two-dimensional views of multiple objects taken from unknown viewpoints. Our approach called ``projective generative adversarial networks" (PrGANs) trains a deep generative model of 3D shapes whose projections match the distributions of the input 2D views. The addition of a projection module allows us to infer the underlying 3D shape distribution without using any 3D, viewpoint information, or annotation during the learning phase. We show that our approach produces 3D shapes of comparable quality to GANs trained on 3D data for a number of shape categories including chairs, airplanes, and cars. Experiments also show that the disentangled representation of 2D shapes into geometry and viewpoint leads to a good generative model of 2D shapes. The key advantage is that our model allows us to predict 3D, viewpoint, and generate novel views from an input image in a completely unsupervised manner.
\end{abstract}

\section{Introduction}\label{s:intro}

\begin{figure}[t]
\centering
\includegraphics[width=0.8\linewidth]{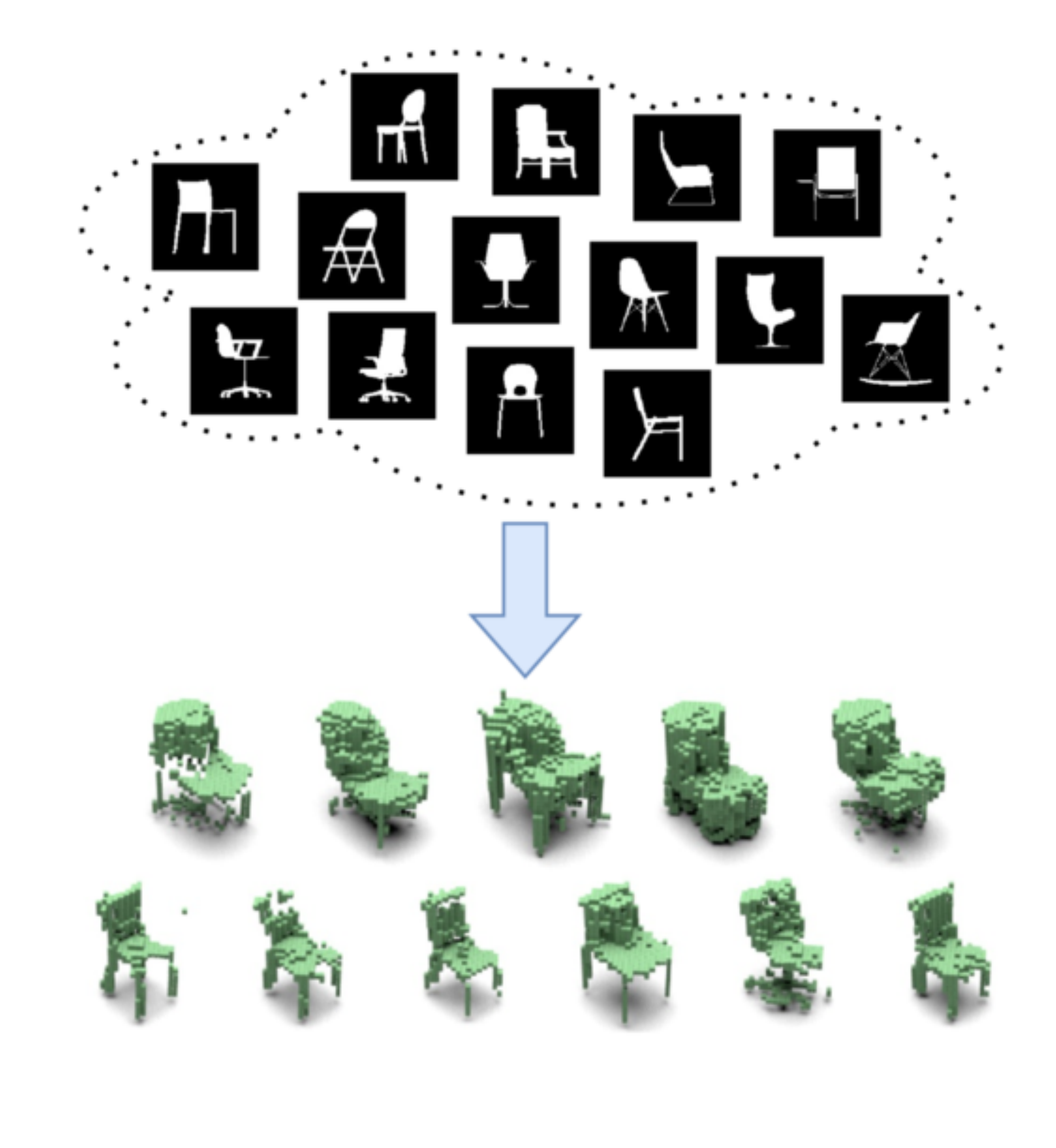}
\caption{\label{f:problem} Given a collection of 2D views of multiple objects, our algorithm infers a generative model of the underlying 3D shapes.}
\vspace{-2pt}
\end{figure}

We live in a three-dimensional (3D) world, but all we see are its projections on to two dimensions (2D). Inferring the 3D shapes of objects from their 2D views is one of the central challenges of computer vision. For example, looking at a catalogue of different chair views in Figure~\ref{f:problem}, one can mentally infer their 3D shapes by simultaneously reasoning about the shared variability in the underlying geometry and viewpoint across instances. In this paper, we investigate the problem of learning a generative model of 3D shapes given a collection of images of an unknown set of objects taken from an unknown set of views.

%Estimating 3D structure from images is a very well-studied area in computer vision. However, there are a number of variants of this problem with varying amounts of difficulty. On the easy end of the spectrum are visual-hull techniques~\cite{} that infer the shape of a particular object given its views from known viewpoints. When the viewpoint is fixed and the lighting is known, photometric stereo~\cite{} can provide accurate geometry estimates for rigid and diffuse surfaces. Next along the spectrum are structure-from-motion techniques~\cite{} that can recover the shape of rigid objects or scenes from their views taken from unknown viewpoints. When the underlying shape and viewpoint both vary, but the shape can be modeled parametrically (e.g., faces), non-rigid structure-from-motion techniques~\cite{} can be applied to simultaneously estimate the 3D shape and viewpoint across image collections. Arguably, the most challenging setting is when a simple generative model of the underlying geometry is not known and only their views are available. Recognition-based techniques for single-image geometry prediction can be applied~\cite{} but these methods rely on strong 3D shape or geometry supervision in order to learn the mapping which may not be readily available.

Although there are several cues for inferring the 3D shape from a single image, in this work we assume that shapes are rendered as binary images bounded by silhouettes.
Even in this simplified setting the problem remains challenging since shading cues are no longer available.
Moreover, which instance was used to generate each image, the viewpoint from which the image was taken, or even the number of underlying instances are not provided. This makes it difficult to apply exisiting approaches of estimating geometry based on structure from motion~\cite{hartley2003multiple,blanz1999morphable}, or computing visual hulls~\cite{laurentini1994visual}.

%there are number of possible 3D shape and viewpoint combinations that give rise to the same image distributions (as seen in Figure~\ref{f:problem}).

The key idea of our approach is to learn a 3D shape generator whose projections match the distribution of the provided images. 
We use the framework of Generative Adversarial Networks~(GANs)~\cite{goodfellow2014generative} and augment the generator with a \emph{projection module}. The generator produces a 3D shape, the projection module renders the shape from a randomly-chosen viewpoint, and the adversarial network discriminates real images from generated ones. 
The projection module is a \emph{differentiable renderer} that approximates the true rendering pipeline and allows us to map 3D shapes to 2D images, as well as back-propagate the gradients of 2D images to 3D shapes. 
Once trained, the model can be used to infer 3D shape distributions from a collection of images (Figure~\ref{f:problem} shows some samples drawn from the generator), and to infer depth data from a single image, without using any 3D or viewpoint information during learning.
We call our approach Projective~GANs~(PrGANs).

There are several challenges that need to be addressed for learning a generative model of 3D shapes from views. 
First is the choice of how the 3D shape is represented. Linear shape basis (or morphable models~\cite{blanz1999morphable,bregler2000recovering}) are effective for categories like faces that have a fixed topology, but less so for categories with varying number of parts, \eg, airplanes and chairs. 
Other bases such as spherical harmonics are not well suited for modeling objects with holes and fine-details. 
Mesh-based representations are commonly used with rendering pipelines (\eg, OpenGL~\cite{woo1999opengl}) and can also be adjusted to match image statistics using a differentiable renderer (\eg, OpenDR~\cite{loper2014opendr}), but the variability of the mesh topology makes it difficult to generate them in a consistent manner across instances.

We make the following assumptions to tackle this problem.
First, shapes are modeled using a voxel representation that indicates the occupancy of a volume in fixed-resolution 3D grid. 
This allows us to model topology in a consistent manner across instances. Second, the effects of lighting and material properties are ignored and a 3D shape is rendered from a given viewpoint as a binary image indicating whether a pixel is occupied or not. 
This allows us to design a volumetric feed-forward network that faithfully reproduces the true rendering pipeline. 
The layers in the feed-forward network are differentiable, allowing the ability to adjust the 3D volume based on projections. 

Our main contributions are as follows: (i) we propose PrGANs, a framework to learn probabilistic distributions over 3D
shapes from a collection of 2D views of objects. We demonstrate its effectiveness on learning complex shape categories
such as chairs, airplanes, and cars sampled from online shape repositories~\cite{chang2015shapenet,wu20153d}. The
results are reasonable, even when views from multiple categories are combined; (ii) On the task of generating 3D shapes,
PrGANs perform well in comparison to GANs trained directly on 3D data~\cite{wu2016learning}; 
(iii) The learned 3D representation can be used for unsupervised estimation of 3D shape and viewpoint given a novel 2D shape, 
and for interpolation between two different shapes.

\section{Related work}\label{s:related}

\paragraph{Estimating 3D shape from image collections.} 
The difficulty of estimating 3D shape can vary widely based on how the images are generated. 
Visual-hull techniques~\cite{laurentini1994visual} can be used to infer the shape of a particular object given its views from known viewpoints by taking the intersection of the projected silhouettes. 
When the viewpoint is fixed and the lighting is known, photometric stereo~\cite{woodham1980photometric} can provide accurate geometry estimates for rigid and diffuse surfaces. 
Structure from motion (SfM)~\cite{hartley2003multiple} can be used to recover the shape of rigid objects from their views taken from unknown viewpoints by jointly reasoning about point correspondences and camera projections. 
Non-rigid SfM can be used to recover shapes from image collections by assuming a parametric family of deformations across 3D shapes.
 An early example of this approach is by Blanz and Vetter~\cite{blanz1999morphable} for estimating 3D shapes of faces from image collections. 
However, 3D data with consistent global correspondences is required in order to learn a morphable model. 
Recently, non-rigid SfM has been applied to categories such as cars and airplanes by manually annotating a fixed set of keypoints across instances in order to bootstrap the learning process~\cite{kar2015category}. 
Our work augments non-rigid SfM using a learned 3D shape generator, which allows us to generalize the technique to categories with diverse structures \emph{without} requiring correspondence annotations. 
Our work is also related to recent work of Kulkarni \etal~\cite{kulkarni2015deep} for estimating a disentangled representation of images into shape, viewpoint, and lighting variables (dubbed ``inverse graphics networks"). However, the shape representation is not explicit, and the approach requires the ability to generate training images while varying one factor at a time.

\paragraph{Inferring 3D shape from a single image.} Optimization-based approaches put priors on geometry, material, and light and estimate all of them by minimizing the reconstruction error when rendered~\cite{land1971lightness,barrow1978recovering,BarronTPAMI2015}.
Recognition-based methods have been used to estimate geometry of outdoor scenes~\cite{hoiem2005geometric,saxena2005learning}, indoor environments~\cite{eigen2015predicting,schwing2012efficient}, and objects~\cite{andriluka2010monocular,savarese20073d}. 
More recently, convolutional networks have been trained to generate views of 3D objects given their attributes and camera parameters~\cite{dosovitskiy2015learning}, to generate 3D shape given a 2D view of the object~\cite{tatarchenko2016multi}, and to generate novel views of an object~\cite{zhou2016view}. Most of these recognition-based methods are trained in a fully-supervised manner and require 3D data, or views of the same object from multiple views, during training.

\begin{figure*}
\centering
\includegraphics[width=0.95\linewidth]{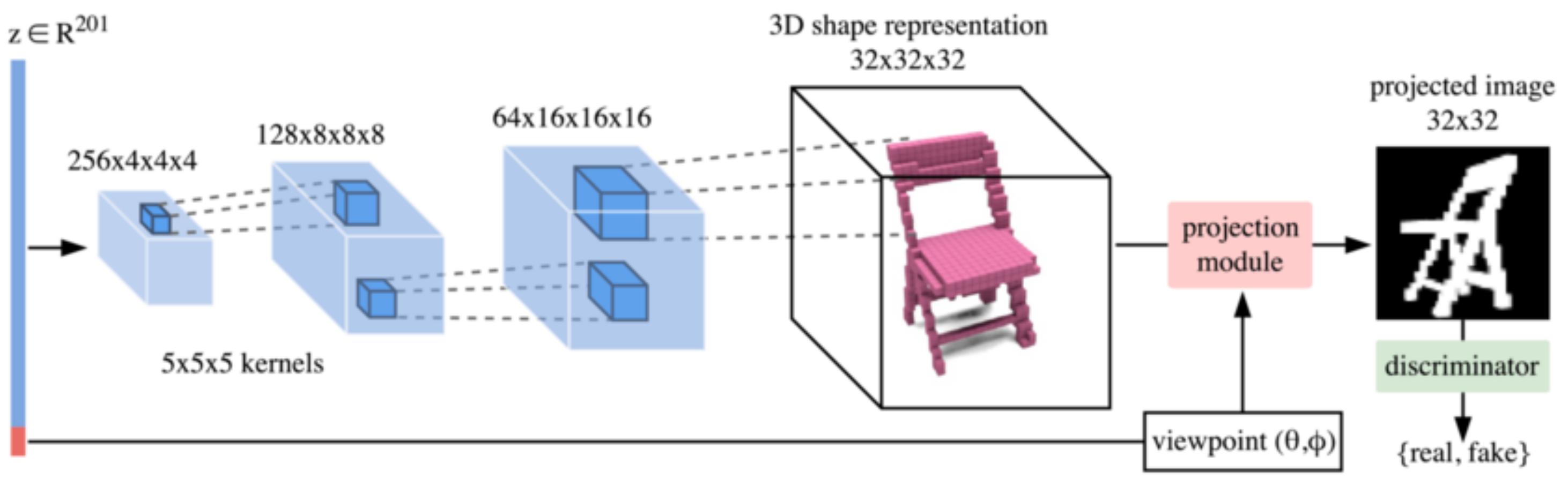}
\caption{\label{fig:prgan-arch}The PrGAN architecture for generating 2D images of shapes. A 3D voxel representation ($32^3$) and viewpoint are independently generated from the input $z$ (201-d vector). The projection module renders the voxel shape from a given viewpoint $(\theta,\phi)$ to create an image. The discriminator consists of 2D convolutional and pooling layers and aims to classify if the input image is generated or real.}
\vspace{-6pt}
\end{figure*}

\paragraph{Generative models for images and shapes.} Our work builds on the success of GANs for generating images across a wide range of domains~\cite{goodfellow2014generative}. Recently, Jiajun \etal~in \cite{wu2016learning} learned a generative model of 3D shapes by using a variant of GAN equipped with 3D convolutions. However, the model was trained using aligned 3D shape data. Our work aims to solve a more difficult question of learning a 3D-GAN from 2D images. Two recent works are in this direction. Rezende \etal in~\cite{rezende2016unsupervised} show results for 3D shape completion for simple shapes when views are provided, but require the viewpoints to be known and the generative models are trained on 3D data. Yan \etal in~\cite{yan2016perspective} learn a mapping from an image to 3D using multiple projections of the 3D shape from known viewpoints (similar to a visual-hull technique). These approaches share some commonalities to ours as the require a way to map 3D shape to 2D images. 

\section{Method}\label{s:method}

Our method builds upon the GAN architecture proposed in Goodfellow~\etal~\cite{goodfellow2014generative}.
The purpose of the GAN is to train a generative model in an adversarial setup.
The model consists of two different parts: a generator and a discriminator.
The generator $G$ aims to transform samples drawn from a simple distribution ${\cal P}$ that appear to have been sampled from the original dataset.
The goal of the discriminator $D$ is to distinguish synthetic samples (created by the generator)
from real samples (drawn from a data distribution ${\cal D}$).
Both the generator and the discriminator are trained jointly to solving for the following optimization:
\begin{equation}\label{eqn:gan}
\min_{G}\max_{D} \mathbb{E}_{x\sim{\cal D}} [ \log \left(D(x)\right) ] + \mathbb{E}_{z\sim{\cal P}} [ \log \left(1-D(G(z))\right) ].
\end{equation}

%\rui{Might be better to present the architecture first, then explain the projection module, training images, and training algorithm.}

Our main task is to train a generative model capable of creating 3D shapes without
relying on 3D data itself, but in 2D images from those shapes, without any view or shape annotation.
In other words, the data distribution consists of 2D images taken from different views and are of different objects.
To address this issue, we factorize the 2D image generator into a 3D shape generator, viewpoint generator, and a projection module (Fig.~\ref{fig:prgan-arch}).
The challenge is to identify a representation suitable for a diverse set of shapes
and a differentiable projection module to create final 2D images and enable end-to-end training. We describe the architecture employed for each of these next. 

\paragraph{3D shape generator.}
The input to the entire generator is $z \in \mathbb{R}^{201}$ with each dimension drawn independently from a uniform distribution $\text{U}(-1,1)$. 
Our 3D shape generator transforms the first 200 dimensions of $z$ to a $32\times32\times32$ voxel representation of the shape.
Each voxel contains a value $v\in[0, 1]$ that represents its occupancy.
The architecture of the 3D shape generator is inspired by the DCGAN~\cite{radford2015unsupervised} and 3D-GAN~\cite{wu2016learning} architectures. 
It consists of a four-layer network shown in Fig.~\ref{fig:prgan-arch}.
The first layer transforms the 200 dimensional vector to a $256\times 4 \times 4 \times 4$ vector using a fully-connected layer.
Subsequent layers have batch normalization and ReLU layers between them
and use 3D kernels of size $5\times5\times5$.
The last layer is succeeded by a sigmoid activation instead of a ReLU.

\begin{figure}
\includegraphics[width=0.95\linewidth]{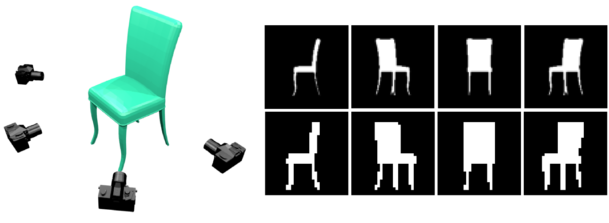}
\caption{\label{fig:projection} Our dataset samples eight viewpoints evenly spaced around the y-axis (only four are shown and the remaining four are diametrically opposite to these). The top four images on the right are training images from the dataset. The bottom four images are created by using the projection module on the voxel representation of the object. Note that the top and bottom match qualitatively.}
\vspace{-8pt}
\end{figure}

\paragraph{Viewpoint generator.}
The viewpoint generator takes the last dimension of $z \in \text{U}(-1,1)$ and transforms it to a viewpoint vector $(\theta,\phi)$.
The training images are assumed to have been generated from 3D models that are upright oriented along a consistent vertical axis (\eg, y-axis), and are centered at the origin.
Most models in online repositories and the real world satisfy this assumption (\eg, chairs are on horizontal planes). We generate images by sampling views uniformly at random from one of eight pre-selected directions evenly spaced around the y-axis (\ie, $\theta=0$ and $\phi=0^\circ$, $45^\circ$, $90^\circ$, $...$, $315^\circ$), as seen in Fig~\ref{fig:projection}.
%passing through the midpoint of the object .
Thus the viewpoint generator picks one of these directions uniformly at random.

\paragraph{Projection module.}

%The projection module may be divided in two separated parts: transformation and sampling
The projection module works as follows.
The first step is to rotate the voxel grid to the corresponding viewpoint.
Let $V : \mathbb{Z}^3 \rightarrow [0,1] \in \mathbb{R}$ be the voxel grid, a function that given
given an integer 3D coordinate $c=(i,j,k)$ returns the occupancy of the voxel centered at $c$.
The rotated version of the voxel grid $V(c)$ is defined as 
$V_{\mathbb{\theta, \phi}} = V(\floor{R(c, \mathbf{\theta,\phi})})$,
where $R(c, \mathbf{\theta},\phi)$ is the coordinate obtained by rotating $c$ around the origin
according to the spherical angles $\mathbf{(\theta,\phi)}$. For simplicity we use nearest neighbor sampling (hence the floor operator).

The second step is to perform the projection to create an image from the rotated voxel
grid.
This is done by applying the projection operator 
%$P : \mathbb{Z}^2, (\mathbb{Z}^3 \rightarrow [0,1]) \rightarrow [0,1]$,
$P((i,j),V) = 1 - e^{-\sum_{k}V(i,j,k)}$. Intuitively, the operator sums up the voxel occupancy values along each line of sight (assuming othographic projection), and applies exponential falloff to create a smooth and differentiable function. When there is no voxel along the line of sight, the value is 0; as the number of voxels increases, the value approaches 1. Combined with the rotated version of the voxel grid, we define our final projection module as:
$P_{\theta,\phi}((i,j),V) = 1 - e^{-\sum_{k}V_{\theta,\phi}(i,j,k)}$. 
As seen in Fig.~\ref{fig:projection} the projection module can well approximate the rendering of a 3D shape as a binary silhouette image, and is differentiable.

\paragraph{Discriminator.}
The discriminator consists of a sequence of 2D convolutional layers with
batch normalization and LeakyReLU layers~\cite{maas2013rectifier} between them. The sequence of transformations done in the network are: $32\times32 \rightarrow 256\times16\times16 \rightarrow 512 \times 8 \times 8 \rightarrow 1024 \times 4 \times 4 \rightarrow 1$.
Similarly to the generator, the last layer of the discriminator is
followed by a sigmoid activation instead of a LeakyReLU.

\paragraph{Training details.}
We train the entire architecture by optimizing the objective in Equation~\ref{eqn:gan}.
Usually, the training updates to minimize each one of the losses is applied once at each
iteration.
However, in our model, the generator and the discriminator have a considerably different
number of parameters.
The generator is trying to create 3D shapes, while the discriminator is trying to
classify 2D images.
To mitigate this issue, we employ an adaptive training strategy.
At each iteration of the training, if the discriminator accuracy is higher than 75\%,
we skip its training.
We also set different different learning rates for the discriminator and the generator:
$10^{-5}$ and $0.0025$, respectively.
Similarly to the DCGAN architecture \cite{radford2015unsupervised}, we use ADAM with $\beta=0.5$ for the optimization.

\section{Experiments}\label{s:experiments}

In this section we describe the set of experiments to evaluate our method
and to demonstrate the extension of its capabilities.
First, we show the effectiveness of our method as both 2D and 3D shape generators.
To this end, we compare our model with a traditional GAN for image generation and a
GAN for 3D shapes.
We present quantitative and qualitative results.
Second, we demonstrate that our method is able to induce 3D shapes from
unlabelled images even when there is only a single view per object.
Third, we present 3D shapes induced by our model from a variety of categories such as airplanes, cars, chairs, motorbikes, and vases. Using the same architecture, we show how our model is able to induce coherent 3D shapes when the training data contains images mixed from multiple categories.
Finally, we show applications of our method in predicting 3D shape from a novel 2D shape, and performing shape interpolation.

\begin{figure}[t]
\setlength{\tabcolsep}{0pt}
\begin{tabular}{cccc}
\includegraphics[width=.25\linewidth]{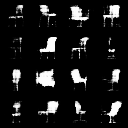} &
\includegraphics[width=.25\linewidth]{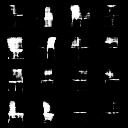} &
\includegraphics[width=.25\linewidth]{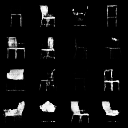} &
\includegraphics[width=.25\linewidth]{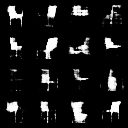} \\
\multicolumn{4}{c}{(a) Results from 2D-GAN.} \vspace{4pt}\\
\includegraphics[width=.25\linewidth]{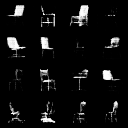} &
\includegraphics[width=.25\linewidth]{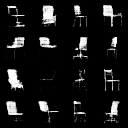} &
\includegraphics[width=.25\linewidth]{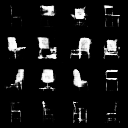} &
\includegraphics[width=.25\linewidth]{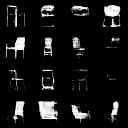} \\
\multicolumn{4}{c}{(a) Results from PrGAN.} \vspace{4pt}
\end{tabular}
\caption{\label{fig:validation2} Comparision between 2D-GAN~\cite{goodfellow2014generative} and our PrGAN model for image generation on the chairs dataset. Refer to Fig.~\ref{fig:generate-shapes} third row, left column for samples of the input data.}
\vspace{-6pt}
\end{figure}

\paragraph{Input data.} We generate training images synthetically using several categories of 3D shapes available in the
ModelNet~\cite{wu20153d} and ShapeNet~\cite{chang2015shapenet} databases. Each category contains a few hundred to thousand shapes. We render each
shape from 8 evenly spaced viewing angles with orthographic projection to produce binary images. To reduce aliasing, we
render each image at $64\times64$ resolution and downsample to $32\times32$. We have found that this generally improves
the results. Using synthetic data allows us to easily perform controlled experiments to analyze our method. 
It is also possible to use real images downloaded from a search engine as discussed in Section \ref{s:discussion}.

\paragraph{Validation.}
We quantitatively evaluate our model by comparing its ability to generate
2D and 3D shapes.
To do so, we use 2D image GAN similar to DCGAN~\cite{radford2015unsupervised} and a 3D-GAN
similar to the one presented at \cite{wu2016learning}.
At the time of this writing the implementation of \cite{wu2016learning} is not public yet, therefore we
implemented our own version.
We will refer to them as 2D-GAN and 3D-GAN, respectively.
The 2D-GAN has the same discriminator architecture as the PrGAN, but the generator contains a sequence of 2D transposed convolutions instead of 3D ones, and
the projection module is removed.
The 3D-GAN has a discriminator with 3D convolutions instead of 3D ones.
The 3D-GAN generator is the same of the PrGAN, but without the projection module.

%Maybe this should be in the method section?
The models used in this experiment are chairs from ModelNet dataset~\cite{wu20153d}.
From those models, we create two sets of training data: voxel grids and images.
The voxel grids are generated by densely sampling the surface and inside of each mesh, and binning the sample points into $32\times32\times32$ 
grid.
A value 1 is assigned to any voxel that contains at least one sample point, and 0 otherwise.
Notice that the voxel grids are only used to train the 3D-GAN, while the images are used to train the 2D-GAN and our PrGAN.
%The images are created using the same models and rendered a binary images.
% Regions that contain the model are white, and regions that do not contain the model are black.
%For each chair model, we create 8 different images by uniformly
%rotating the model around the vertical axis.
%Also, we observed that the performances of the 2D-GAN and the PrGAN are enhanced if the images
%are rendered in $64\times64$ and downsampled to $32\times32$.
%Notice that the data used to train the PrGAN consists of images, but the method is able to learn a generative model for 3D shapes from those images, which makes it comparable to the 3D-GAN.

\begin{figure}[t]
\setlength{\tabcolsep}{0pt}
\begin{tabular}{cccc}
\includegraphics[width=.24\linewidth]{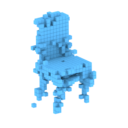} &
\includegraphics[width=.24\linewidth]{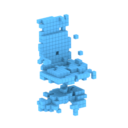} &
\includegraphics[width=.24\linewidth]{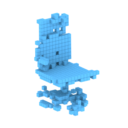} &
\includegraphics[width=.24\linewidth]{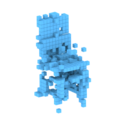} \\
\includegraphics[width=.24\linewidth]{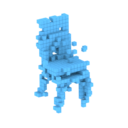} &
\includegraphics[width=.24\linewidth]{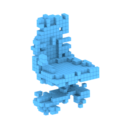} &
\includegraphics[width=.24\linewidth]{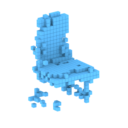} &
\includegraphics[width=.24\linewidth]{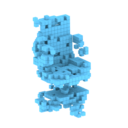} \\
\multicolumn{4}{c}{(a) Results from 3D-GAN.} \vspace{4pt}\\
\includegraphics[width=.24\linewidth]{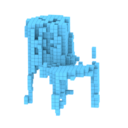} &
\includegraphics[width=.24\linewidth]{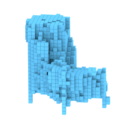} &
\includegraphics[width=.24\linewidth]{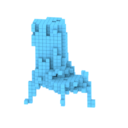} &
\includegraphics[width=.24\linewidth]{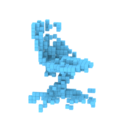} \\
\includegraphics[width=.24\linewidth]{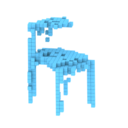} &
\includegraphics[width=.24\linewidth]{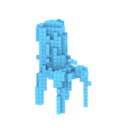} &
\includegraphics[width=.24\linewidth]{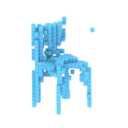} &
\includegraphics[width=.24\linewidth]{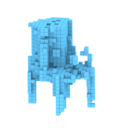} \\
\multicolumn{4}{c}{(a) Results from PrGAN.} \vspace{4pt}\\
\end{tabular}
\caption{\label{fig:validation3} Comparison between 3D-GAN~\cite{wu2016learning} and our PrGAN for 3D shape generation. The 3D-GAN is trained on 3D voxel representation of the chair models, and the PrGAN is trained on images of the chair models (refer to Fig.~\ref{fig:generate-shapes} third row).}
\vspace{-8pt}
\end{figure}

Our quantitative evaluation is done by taking the Maximum Mean Discrepancy (MMD)~\cite{gretton2006kernel} 
between the data created by the generative models and the training data.
We use a kernel bandwidth of $10^{-3}$ for images and $10^{-2}$ for voxel grids.
The training data consists of 989 voxel grids and 7912 images.
To compute the MMD, we draw 128 random data points from each one of the generative models.
The distance metric between the data points is the hamming distance divided by the dimensionality of the data.
Because the data represents continuous occupancy values, we binaritize them by using a threshold of $0.001$ for images or voxels created
by PrGAN, and $0.1$ for voxels created by the 3D-GAN.

\begin{figure}[t]
\includegraphics[width=0.95\linewidth]{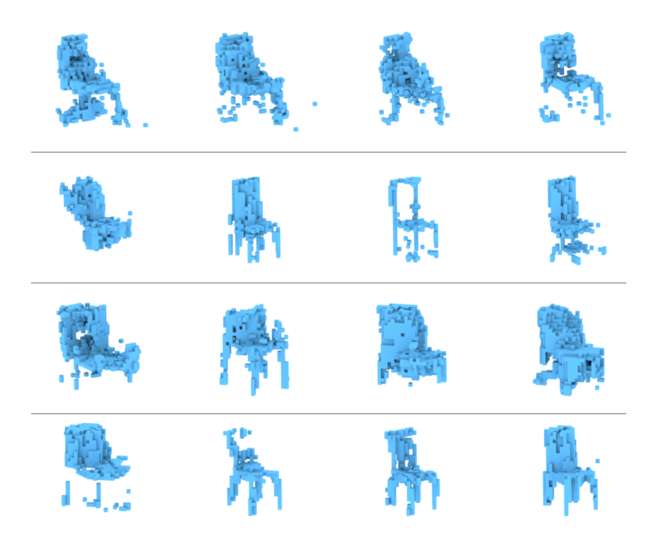}
\caption{\label{fig:varying_views} Shapes generated from PrGAN by varying the number of views per object in the training data. From the top row to the bottom row, the number of views per object in the training set are 1, 2, 4, and 8 respectively.}
\vspace{-8pt}
\end{figure}

Results show that for 2D-GAN, the MMD between the generated images and the training data is 90.13. For PrGAN, the MMD is 88.31, which is slightly better quantitatively than 2D-GAN. Fig.~\ref{fig:validation2} shows a qualitative comparison. The results are visually very similar. For 3D-GAN, the MMD between the generated voxel grids and the training voxel grids is 347.55. For PrGAN, the MMD is 442.98, which is worse compared to 3D-GAN. This is not surprising as 3D-GAN is trained on 3D data, while PrGAN is trained on the image views only. Fig.~\ref{fig:validation3} presents a qualitative comparison. In general PrGAN has trouble generating intrior structures because the training images are binary, carries no shading information, and are taken from a limited set of viewing angles. Nonetheless, it learns to generate exterior structures reasonably well.

\paragraph{Varying the number of views per model.} In the default setting, our training data consists of 8 views per
object. Here we study the ability of our method in the more challenging case where the training data contains fewer
number of views per object. To do so, we generate a new training set that contains only 1 randomly chosen view
(among the 8) per object and use it to train PrGAN. We then repeat the experiments for 2 randomly chosen views per object, and also 4. The
results are shown in Fig.~\ref{fig:varying_views}. The 3D shapes that PrGAN learns becomes increasingly better as the
number of views increases. Even in the single view per object case, our method is able to induce reasonable shapes. h

%Our standard experimental setup uses eight images taken from the same eight different viewpoints.
%An interesting question is to assess the ability of the method to learn
%how to generate shapes when the one or more of the views is not accessible.
%To evaluate this issue, we create three different datasets.
%The first one, contains only one randomly selected image from the eight previously available.
%The second and third dataset contain two and four images.
%As we can see in Fig.~\ref{fig:varying_views}, our method is able to induce
%reasonable shapes even when a single view is available.
%However, the quality improves as the number of available views increases.

\begin{figure*}
  \newcommand{\fh}{0.19\linewidth}
  \begin{center}
  \setlength{\tabcolsep}{3pt}
  \begin{tabular}{ccc}
    \textbf{Input} & \textbf{Generated images} & \textbf{Generated shapes} \\
    \includegraphics[height=\fh]{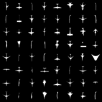} & 
    \includegraphics[height=\fh]{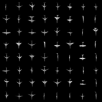} & 
    \includegraphics[height=\fh]{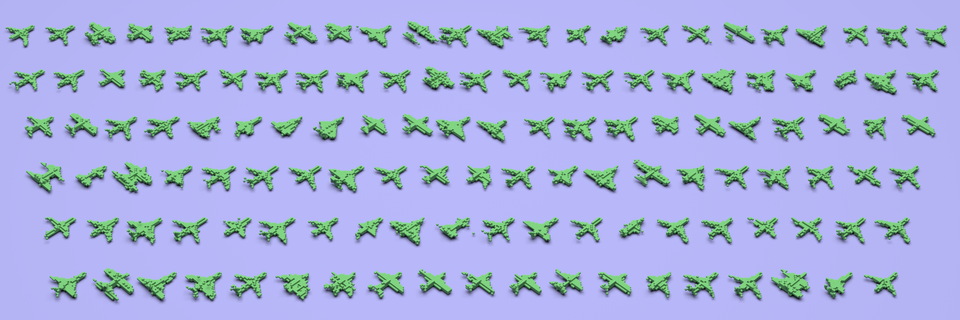} \\
    \includegraphics[height=\fh]{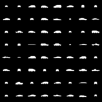} & 
    \includegraphics[height=\fh]{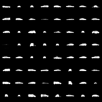} & 
    \includegraphics[height=\fh]{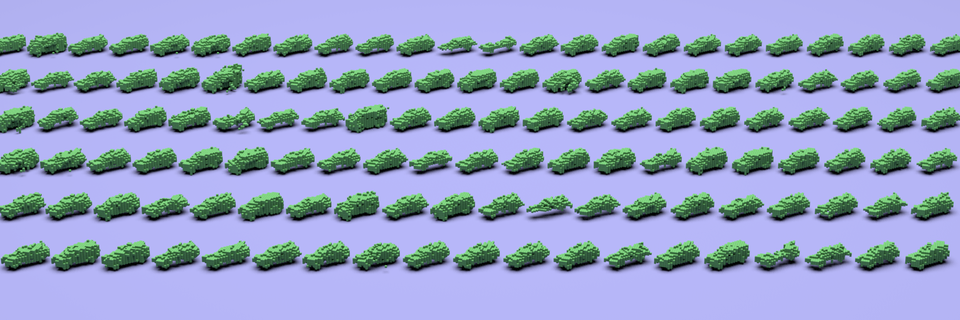} \\
    \includegraphics[height=\fh]{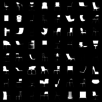} & 
    \includegraphics[height=\fh]{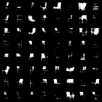} & 
    \includegraphics[height=\fh]{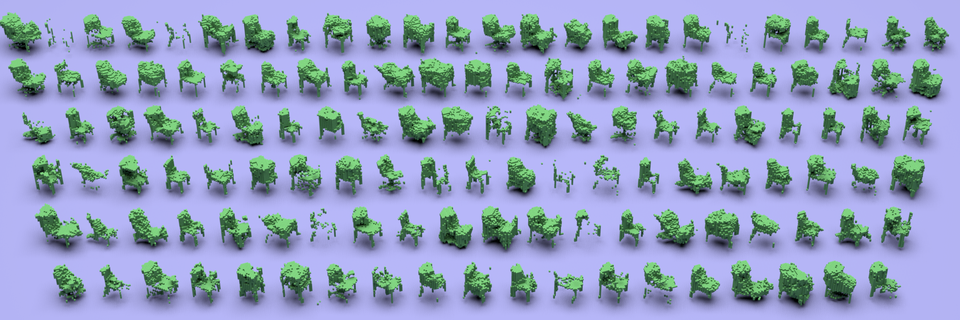} \\
    \includegraphics[height=\fh]{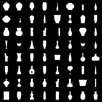} & 
    \includegraphics[height=\fh]{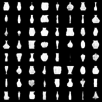} & 
    \includegraphics[height=\fh]{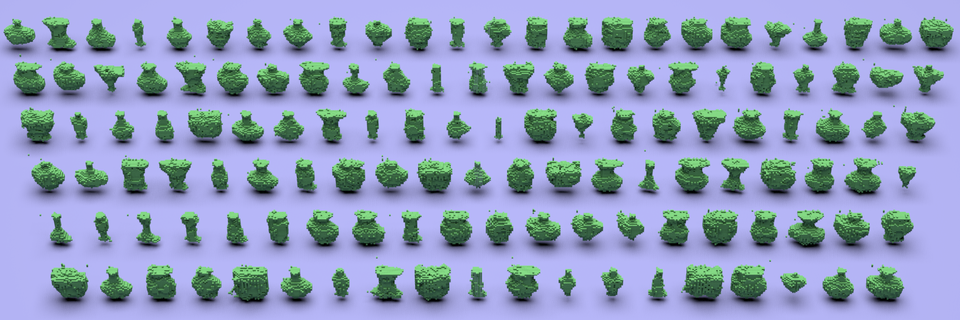} \\
    \includegraphics[height=\fh]{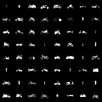} & 
    \includegraphics[height=\fh]{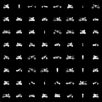} & 
    \includegraphics[height=\fh]{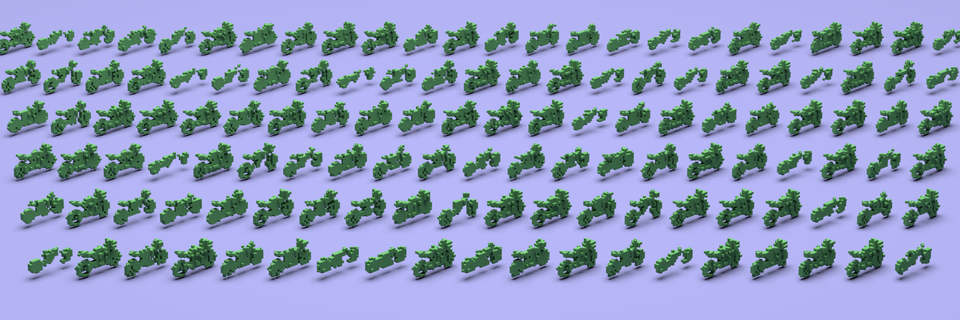} \\
    \includegraphics[height=\fh]{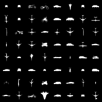} & 
    \includegraphics[height=\fh]{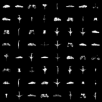} & 
    \includegraphics[height=\fh]{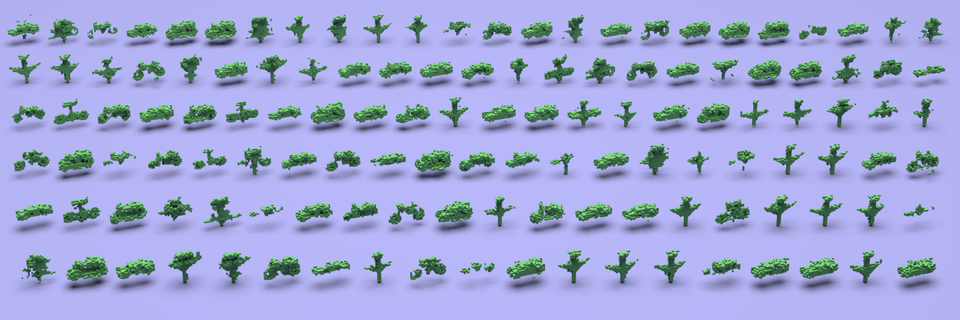} \\
  \end{tabular}
  \end{center}
  \vspace{-6pt}
  \caption{\label{fig:generate-shapes} Results for 3D shape induction using PrGANs. From top to bottom we show results for airplane, car, chair, vase, motorbike, and a 'mixed' category obtained by combining training images from airplane, car, and motorbike. At each row, we show on the left 64 randomly sampled images from the input data to the algorithm, on the right 128 sampled 3D shapes from PrGAN, and in the middle 64 sampled images after the projection module is applied to the generated 3D shapes. The model is able to induce a rich 3D shape distribution for each category. The mixed-category produces reasonable 3D shapes across all three combined categories. Zoom in to see details.}
\end{figure*}

\begin{figure*}[t]
\centering
\setlength{\tabcolsep}{0pt}
\begin{tabular}{cccccccccc}
\includegraphics[width=.1\linewidth]{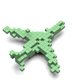} &
\includegraphics[width=.1\linewidth]{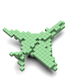} &
\includegraphics[width=.1\linewidth]{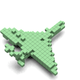} &
\includegraphics[width=.1\linewidth]{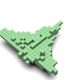} &
\includegraphics[width=.1\linewidth]{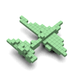} &
\includegraphics[width=.1\linewidth]{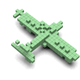} &
\includegraphics[width=.1\linewidth]{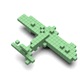} &
\includegraphics[width=.1\linewidth]{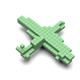} &
\includegraphics[width=.1\linewidth]{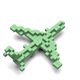} &
\includegraphics[width=.1\linewidth]{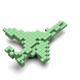} \\
\includegraphics[width=.1\linewidth]{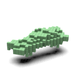} &
\includegraphics[width=.1\linewidth]{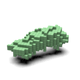} &
\includegraphics[width=.1\linewidth]{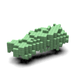} &
\includegraphics[width=.1\linewidth]{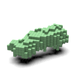} &
\includegraphics[width=.1\linewidth]{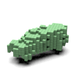} &
\includegraphics[width=.1\linewidth]{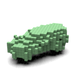} &
\includegraphics[width=.1\linewidth]{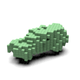} &
\includegraphics[width=.1\linewidth]{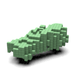} &
\includegraphics[width=.1\linewidth]{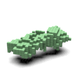} &
\includegraphics[width=.1\linewidth]{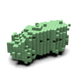} \\
\includegraphics[width=.1\linewidth]{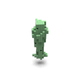} &
\includegraphics[width=.1\linewidth]{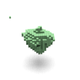} &
\includegraphics[width=.1\linewidth]{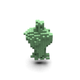} &
\includegraphics[width=.1\linewidth]{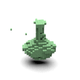} &
\includegraphics[width=.1\linewidth]{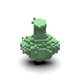} &
\includegraphics[width=.1\linewidth]{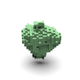} &
\includegraphics[width=.1\linewidth]{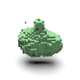} &
\includegraphics[width=.1\linewidth]{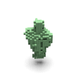} &
\includegraphics[width=.1\linewidth]{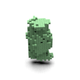} &
\includegraphics[width=.1\linewidth]{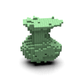} \\
\includegraphics[width=.1\linewidth]{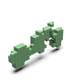} &
\includegraphics[width=.1\linewidth]{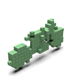} &
\includegraphics[width=.1\linewidth]{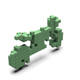} &
\includegraphics[width=.1\linewidth]{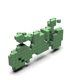} &
\includegraphics[width=.1\linewidth]{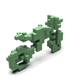} &
\includegraphics[width=.1\linewidth]{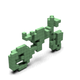} &
\includegraphics[width=.1\linewidth]{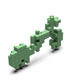} &
\includegraphics[width=.1\linewidth]{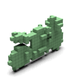} &
\includegraphics[width=.1\linewidth]{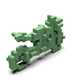} &
\includegraphics[width=.1\linewidth]{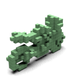} 
\end{tabular}
\caption{\label{fig:example-samples} A variety of 3D shapes generated by  PrGAN trained on 2D views of (from the top row to the bottom row) airplanes, cars, vases, and bikes. These examples are chosen from the gallery in Fig.~\ref{fig:generate-shapes} and demonstrate the quality and diversity of the generated shapes.}\end{figure*}

\paragraph{Visualizations across categories.} Our method is able to generate 3D shapes for a wide range of categories. Fig.~\ref{fig:generate-shapes} show a gallery of results, including airplanes, car, chairs, vases, motorbikes. For each category we show 64 randomly sampled training images, 64 generated images from PrGAN, and renderings of 128 generated 3D shapes (produced by randomly sampling the 200-d input vector of the generator). 
%shows the input images and sampled models and views of airplanes, chairs, vases, and motorbikes.
One remarkable property is that the generator produces 3D shapes in a consistent horizontal and vertical axes, even though the training data is only consistently oriented along the vertical axis. Our hypothesis for this is that the generator finds it more efficient to generate shapes in a consistent manner by sharing parts across models. Fig.~\ref{fig:example-samples} shows selected examples from Fig.~\ref{fig:generate-shapes} that demonstrates the quality and diversity of the generated shapes. 

The last row in Fig.~\ref{fig:generate-shapes} shows an example of a 'mixed' category, where the training images combine the three categories of airplane, car, and motorbike. The same PrGAN network is used to learn the shape distributions. Results show that PrGAN learns to represent all three categories well, without any additional supervision.

%Here we investigate if a single generative model can generate shapes across multiple categories without any additional supervision. For this experiment we combine views of models across airplanes, motorbikes, and cars and provide them as input to the learning algorithm. The bottom row in Figure~\ref{fig:generate-shapes} shows the results. 

\begin{figure}[t]
\setlength{\tabcolsep}{0pt}
\begin{tabular}{cccccc}
\includegraphics[width=.166\linewidth]{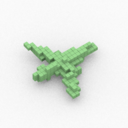} &
\includegraphics[width=.166\linewidth]{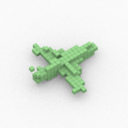} &
\includegraphics[width=.166\linewidth]{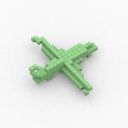} &
\includegraphics[width=.166\linewidth]{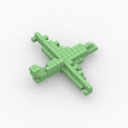} &
\includegraphics[width=.166\linewidth]{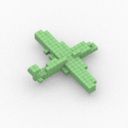} &
\includegraphics[width=.166\linewidth]{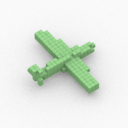} \\
\includegraphics[width=.166\linewidth]{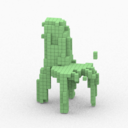} &
\includegraphics[width=.166\linewidth]{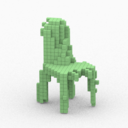} &
\includegraphics[width=.166\linewidth]{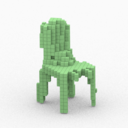} &
\includegraphics[width=.166\linewidth]{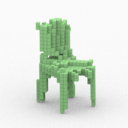} &
\includegraphics[width=.166\linewidth]{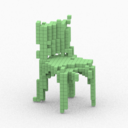} &
\includegraphics[width=.166\linewidth]{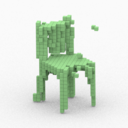} \\
\end{tabular}
\caption{\label{fig:interp} Shape interpolation by linearly interpolating the encodings of the starting shape and ending shape.}
\vspace{-12pt}
\end{figure}

\paragraph{Shape interpolation.}
Once the generator is trained, any encoding $z$ supposedly generates a plausible 3D shape, hence $z$ represents a 3D shape manifold. Similar to previous work, we can interpolate between 3D shapes by linearly interpolating their $z$ codes. Fig.~\ref{fig:interp} shows the interpolation results for two airplane models and two chair models. 
%Such properties have been demonstrated for other deep generative models and we show that
%the generator learned by our method works in the same way.
%Our experiment consists of training our model with the airplane dataset from ShapeNet~\cite{}, selecting
%the encodings of two different shapes and interpolating them.
%As we can see in 

\begin{figure}[t]
\includegraphics[width=0.95\linewidth]{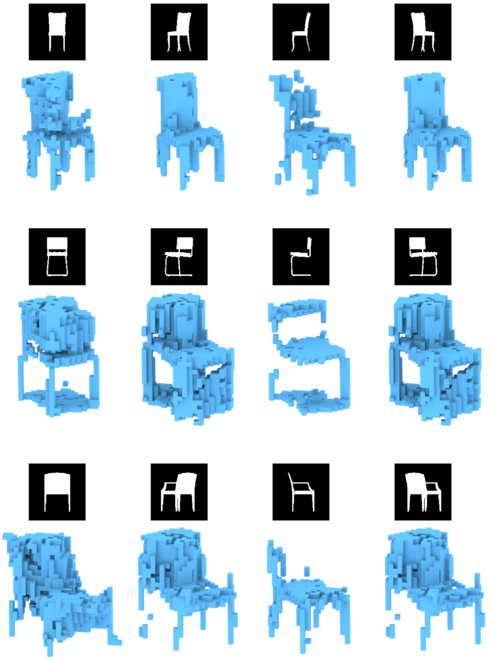}
\caption{\label{fig:sp} Shape infered by a single view image using the encoding network. At each row, the four images are different views of the same chair. The inferred shapes are different, but are nonetheless plausible given the single view.}
\vspace{-8pt}
\end{figure}

\paragraph{Unsupervised shape and viewpoint prediction.}
Our method is also able to handle unsupervised prediction of shapes in 2D images.
Once trained, the 3D shape generator is capable of creating shapes from
a set of encodings $z \in \mathbb{R}^{201}$.
One application is to predict the encoding of the underlying 3D object given a single view image of the object.
We do so by using the PrGAN's generator to produce a large number of encoding-image pairs, then use the data to train a neural network (called encoding network). In other words, we create a training set that consists of images synthesized by the PrGAN and the encodings that generated them. The encoding network is fully connected, with 2 hidden layers, each with 512 neurons.
The input of the network is an image and the output is an encoding.
The last dimension of $z$ describes the view, and the first 200 dimensions describe the code of the shape, which allows us to further reconstruct the 3D shape as a $32^3$ voxel grid. With the encoding network, we can present to it a single view image, and it outputs the shape code along with the viewing angle. 
%We use these 200 values to retrieve the shape present in the image.
%The network is trained with images created by the PrGAN, but we present the results of the network
Experimental results are shown in in Figure~\ref{fig:sp}. This whole process constitutes a completely unsupervised approach to creating a model that infers a 3D shape from a single image.

\section{Limitations and Future Work}\label{s:discussion}
\begin{figure}[t]
\includegraphics[width=0.95\linewidth]{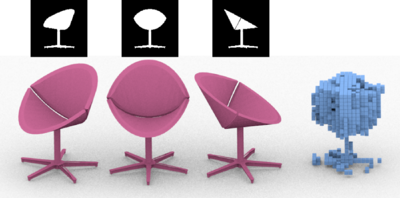}
\caption{\label{fig:failure} Our method is unable to capture the concave interior structures in this chair shape. The pink shapes show the original shape used to generate the projected training data, shown by the three binary images on the top (in high resolution). The blue voxel representation is the inferred shape by our model. Notice the lack of internal structure.}
\end{figure}

\paragraph{Failure cases.} Comprared to 3D-GANs, the proposed PrGAN models cannot discover structures that are hidden
due to occlusions from all views. For example, it fails to discover that some chairs have concave interiors and the generator simply fills these since it does not change the silhouette from any view as we can see at Figure
\ref{fig:failure}. 
However, this is a natural drawback of view-based approaches since some 3D ambiquities cannot be resolved (\eg necker cubes) without relying on other cues. Despite this, one advantage over 3D-GAN is that our model does not require consistently aligned 3D shapes since it reasons over viewpoints.

\paragraph{Higher-resolution models.} Another drawback of our approach is that we currently generate low-resolution ($32^3$) shapes. This is an inherent limitation of voxel-based representations since the size of the voxel grid scales cubically with the resolution. Recent results in learning generative models of images using residual architectures~\cite{he2015deep} and multi-scale reasoning~\cite{denton2015deep}, may help scale the resolution of generative models to the next few octaves.

\paragraph{Using multiple cues for shape reasoning.} Our approach currently only relies on binary silhouettes for estimating the shape. This contributes to the lack of geometric details, which can be improved by incorporating shading cues. One strategy is to train a more powerful differentiable function approximator, \eg, a convolutional network, to replicate the sophisticated rendering pipelines developed by the computer graphics community. 
Once trained, the resulting \emph{neural renderer} could be a plug-in replacement for the \emph{projection module} in the PrGAN framework.
This would allow the ability to use collections of realistically-shaded images for infering probabilistic models of 3D shapes and other properties.
Recent work on sceen-space shading using convnets are promising~\cite{nalbach2016deep}.

\paragraph{Learning from real images.} Our approach can be extended to learning 3D shapes from real images by applying an exisiting approach for segmentation such as~\cite{long2015fully}. However, the assumption that the viewpoints are uniformly distributed over the viewing sphere may not hold. In this situation, one can either learn a distribution over viewpoints by mapping a few dimensions of the input code $z$ to a distribution over viewpoints $(\theta,\phi)$ using a multi-layer network. More generally, one can also learn a distribution over a full set of camera parameters. An alternative is learn a conditional generator where the viewpoint is provided as input to the algorithm. This may be obtained using a generic viewpoint estimator such as~\cite{tulsiani2015pose,su2015render}. We will explore these directions in future work.

%We have conducted an initial experiment in this direction. The dataset consists of views of cubiods with different colors on each side of the cuboid. In this dataset each shape can be completely represented by vector $z \in \mathbb{R}^{21}$ corresponding to the width, height, depth, and (R,G,B) colors for each of the six faces. We train a neural renderer that learn a mapping from $z, (\theta, \phi)$ to an image by synthetically generating cuboids with different colored faces and rendering them in OpenGL. The neural renderer architecture looks very similar to a image generator in DCGAN, but can be trained in a fully-supervised manner since there is practically unlimited training data. We found that we were able to train a neural renderer to produce $64\times64$ images accurately with $XXXX$ training images. \smaji{describe dataset, show examples from the 2DGAN and PrGAN}.

\paragraph{Conclusion.} In summary, we have proposed a framework for infering 3D shape distributions from 2D shape collections by agumenting a convnet-based 3D shape generator with a projection module. This compliments exisiting approches for non-rigid SfM since these models can be trained without prior knowledge about the shape family, and can generalize to categories with variable structure. We showed that our models can infer 3D shapes for a wide range of categories, and can be used to infer shape and viewpoint from a single image in a completely unsupervised manner. We believe that the idea of using a differentiable render to infer distributions over unobserved scene properties from images can be applied to other problems.

\clearpage
{\small
\bibliographystyle{ieee}
\bibliography{egbib}
}

\end{document}